\documentclass[10pt,twocolumn,letterpaper]{article}

\usepackage{cvpr}              

\usepackage{amsmath}
\usepackage{listings}
\usepackage{relsize}
\usepackage{algorithm}
\usepackage{algorithmic}
\usepackage{xcolor}

\lstdefinelanguage{json}{
    basicstyle=\ttfamily\footnotesize,
    breaklines=true,
    stringstyle=\color{red},
    showstringspaces=false,
    keywordstyle=\color{blue}\bfseries,
    commentstyle=\color{gray}\itshape,
    frame=single,
    morestring=[b]",
    morecomment=[l]{//},
    morekeywords={true, false, null}
}

\newcommand{\mymethod}[1]{City$\mathcal{X}$}

\definecolor{cvprblue}{rgb}{0.21,0.49,0.74}
\usepackage[pagebackref,breaklinks,colorlinks,allcolors=cvprblue]{hyperref}

\title{\mymethod{}: Controllable Procedural Content Generation for Unbounded 3D Cities}

\makeatletter
\newcommand{\printfnsymbol}[1]{\textsuperscript{\@fnsymbol{#1}}}
\makeatother

\author{
Shougao Zhang$^{2}$\thanks{Equal contributions.} \and
Mengqi Zhou$^{1}$\printfnsymbol{1} \and
Yuxi Wang$^{5}$\printfnsymbol{1} \and
Chuanchen Luo$^{4}$ \and
Rongyu Wang$^{1}$ \and
Yiwei Li$^{1}$ \and
Zhaoxiang Zhang$^{1,3,5}$\thanks{Corresponding author} \and
Junran Peng$^{1,3,6}\footnotemark[2]$
}

\begin{document}

\twocolumn[{%
\renewcommand\twocolumn[1][]{#1}%
\maketitle

\vspace{-3.5em}
\begin{center}
\small
$^1$University of Chinese Academy of Sciences, $^2$China University of Geosciences (Beijing) \\
$^3$Institute of Automation, Chinese Academy of Sciences (CASIA), $^4$Shandong University \\
$^5$Centre for Artificial Intelligence and Robotics,(HKISI\_CAS), $^6$University of Science and Technology Beijing\\
\texttt{zhangshougao@email.cugb.edu.cn}, \texttt{jrpeng4ever@126.com}, \texttt{zhoumengqi2022@ia.ac.cn} \\
\texttt{yuxi.wang93@gmail.com}, \texttt{chuanchen.luo@sdu.edu.cn},
\texttt{zhaoxiang.zhang@ia.ac.cn}
\end{center}

\includegraphics[width=1\linewidth]{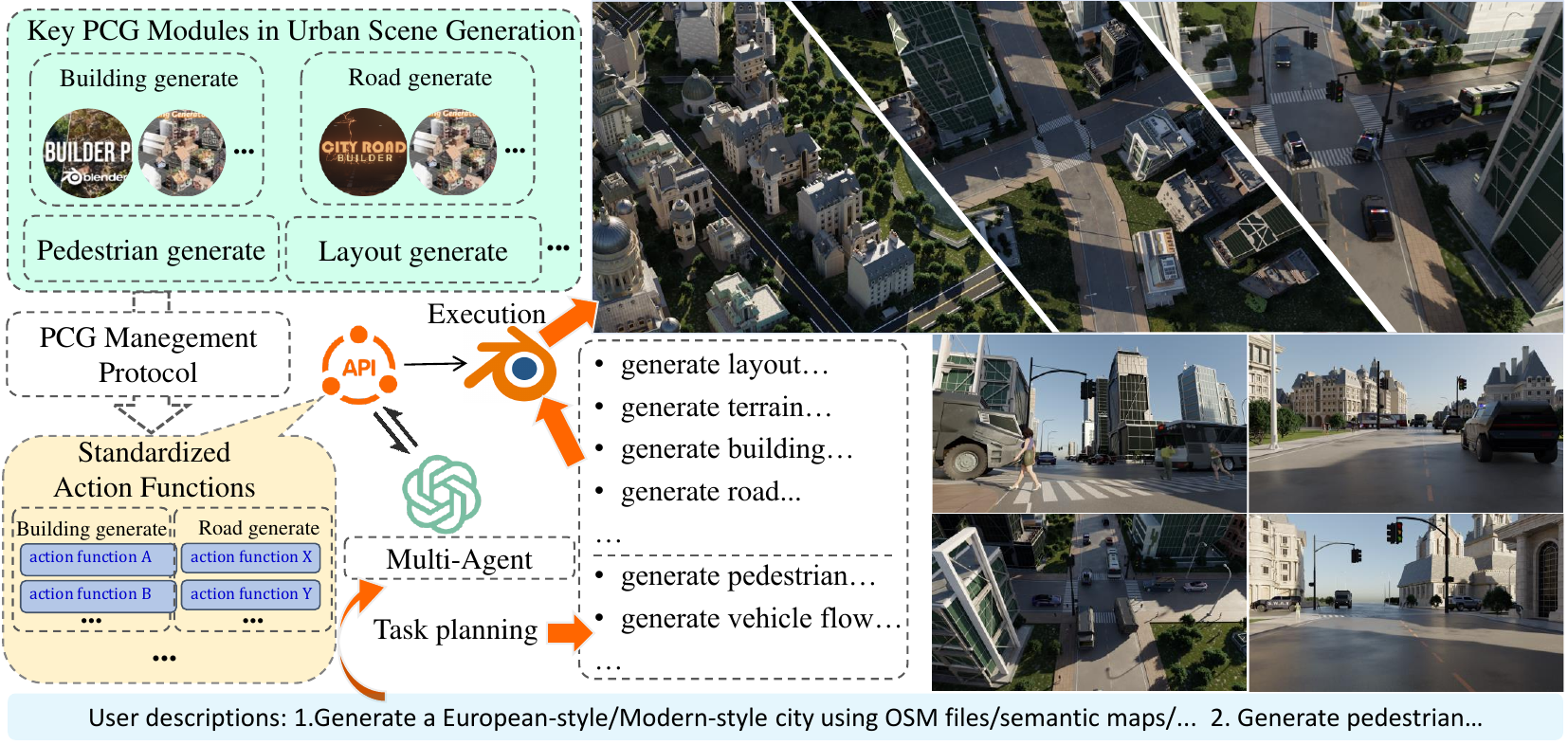}
\captionof{figure}{The proposed CityX can generate high-quality, controllable, and editable large-scale urban scenes based on user descriptions and multimodal inputs (OSM files, semantic maps, satellite images). The generated scenes allow for the integration of dynamic elements, such as pedestrians and traffic, ensuring a fully interactive and adaptable environment.}
\label{fig:startfig}
\vspace{0.5em}
}]
\vspace{-0.5em}
\begin{abstract}

Urban areas, as the primary human habitat in modern civilization, accommodate a broad spectrum of social activities. 
With the surge of embodied intelligence, recent years have witnessed an increasing presence of physical agents in urban areas, such as autonomous vehicles and delivery robots.
As a result, practitioners significantly value crafting authentic, simulation-ready 3D cities to facilitate the training and verification of such agents.
However, this task is quite challenging.
Current generative methods fall short in either diversity, controllability, or fidelity.
In this work, we resort to the procedural content generation (PCG) technique for high-fidelity generation. 
It assembles superior assets according to empirical rules, ultimately leading to industrial-grade outcomes.
To ensure diverse and self-contained creation, we design a management protocol to accommodate extensive PCG plugins with distinct functions and interfaces. 
Based on this unified PCG library, we develop a multi-agent framework to transform multi-modal instructions, including OSM, semantic maps, and satellite images, into executable programs. 
The programs coordinate relevant plugins to construct the 3D city consistent with the control condition.
A visual feedback scheme is introduced to further refine the initial outcomes.
Our method, named \mymethod, demonstrates its superiority in creating diverse, controllable, and realistic 3D urban scenes.
The synthetic scenes can be seamlessly deployed as a real-time simulator and an infinite data generator for embodied intelligence research. Our project page: \url{https://cityx-lab.github.io/}.

\end{abstract}    
\section{Introduction}

With the development of civilization, cities have become the primary human habitat, supporting a wide range of social activities.
It is characterized by a complex layout, diverse buildings, interlaced roads, and esthetic streetscape.
Due to rapid advancements in robotics and embodied AI, there has been a notable increase in the presence of physical agents in urban areas, including autonomous taxis, unmanned aerial vehicles (UAVs), and mobile robots designed for delivery, cleaning, and service. 
These intelligent agents not only enhance the convenience of daily life for residents but also boost the economic efficiency of society.
By providing a secure and interactive enviroment for training and verfication, urban simulator greatly facilitates the development of such agents.
As a critical component of the urban simulator, 3D city creation has attracted significant attention from both industry and academy.
The simulator has demands a high degree of diversity, controllability, and fidelity in synthetic urban scenes.
These properties are necessary prerequisite for robots to generalize to real-world scenarios.
Such a task is particularly challenging.
A senior designer generally spend a few days crafting a desirable city scene from scratch.

Recently, some methods \cite{InfiniCity,CityGen,xie2023citydreamer} have emerged to automate the creation of 3D urban scenes.
A mainstream practice is to lift 2D generative representation to 3D space.
For instance, InfiniCity \cite{InfiniCity} and CityDreamer \cite{xie2023citydreamer} condition neural radiance field on the bird's-eye-view layout feature map.
Qualitative results show that these methods suffer from fuzzy appearance or irregular geometry.
Moreover, the implicit neural scene representation they used is not compatible with existing simulation engines.
These drawbacks prevent them from practical application.
Realizing the significant gap between the strict demands and the performance of neural methods, researchers have attempted to borrow inspiration from mature industrial pipelines recently.
Procedural content generation (PCG), as a key technique in video game development, is frequently employed for the rapid creation of exquisite scenes. 
Each PCG plugin specializes in creating a specific category.
It assembles desired assets algorithmically based on the empirical rules which are derived by profound analysis in the patterns of instances belonging to this category.
The PCG synthetic assets fully meet the industrial requirements and are seamlessly compatible with simulation engines.
3D-GPT \cite{3D-GPT}, Scenecraft \cite{hu2024scenecraft}, and SceneX \cite{SceneX} are three pioneer methods that resort to the PCG paradigm for scene generation.
They leverage the powerful planning capability of large language models (LLMs) to orchestrate PCG programs according to text instructions.
Despite their favorable performance against those neural methods, these methods are initially designed for the creation of general scenes and have not been tailored for urban scenes  
They only support the control of textual instructions which is too coarse for the 3D urban creation.
While in the industrial pipeline, the designers need to arrange relevant assets, such as buildings, streets, and bridges, according to the layout constraints which are generally defined by OpenStreetMap or semantic maps.
Moreover, current solutions are deficient in adaptability and scalability. 
They lack the ability to coordinate PCG plugins with distinct functions and incompatible interfaces, which in turn impedes the creation of self-contained urban scenes.

In this paper, we introduce an efficient and productive pipeline, named \mymethod{}, for the controllable creation of diverse and high-fidelity 3D cities.
To tackle challenges in the integration of various plugins, we first develop a management protocol to reformat the interface of different assets. 
The protocol primarily comprises two parts: 
(i) a dynamic API conversion interface, enabling free and flexible integration of plugins with different functions; 
(ii) structured encapsulation, formatting the plugins for subsequent calling.
The proposed protocol shows the potential to build an ecosystem for the community, which is crucial for unifying various PCG plugins for the integral creation of complex scenes. 
Based on this unified protocol, we design a LLM multi-agent framework to automate the orchestration of PCG assets.
The framework accepts text descriptions, XML-formatted OSM data, semantic maps, and satellite images as control instructions.
The LLM agents perform task planning, plan validation, and action execution sequentially to transform instructions into executable Blender programs.
Afterward, Blender interprets the program and renders the scene.
A visual language model (VLM) is then introduced to inspect the creation from rendering and offer feedback to refine the initial results further.

Our synthetic 3D cities excel in controllability, diversity, and fidelity. 
For completeness, we have employed another multi-agent framework to incorporate dynamic elements, such as vehicles and pedestrians, into the synthetic scenes.
These dynamic elements navigate and move within their designated feasible areas. 
Both static and dynamic assets in the scene are interactive, which is essential for tasks involving grasping, manipulation, and active obstacle avoidance. Overall, our synthesized scenes are well-suited to serve as the foundational environment for the simulator.

The contributions of this paper are summarized as follows:
\begin{itemize}
    \item We propose \mymethod{}, an effective and efficient pipeline to create diverse, controllable, and realistic 3D city scenes. Our synthetic scenes are ready for the simulation of embodied AI, facilitating the training and deployment of physical agents.
    \item We design a management protocol to enable the unified calling of PCG plugins with distinct interfaces, laying the foundation for further extension.
    \item We develop an LLM-based multi-agent framework to translate multi-modal instructions into executable blender programs. It coordinates diverse PCG plugins to create harmonized scenes consistent with control conditions.
\end{itemize}

\section{Related Works}
Generating a 3D urban scene is a complex task involving multiple modules, such as accurately generating a reasonable layout and then constructing appropriate instances on the layout. Additionally, applying Multi-Agent systems to complex tasks like 3D urban scene generation presents significant challenges. In this section, we will discuss works related to these aspects.

\noindent\textbf{Agent Systems Based on LLMs.} When researching agent systems based on Large Language Models (LLMs), the focus lies on effectively integrating and applying these models to execute complex tasks. Existing relevant work encompasses various aspects, including task management, role-playing, dialogue patterns, and tool integration. For example, \cite{huang2022language} utilizes the expansive domain knowledge of LLMs on the internet and their emerging zero-shot planning capabilities to execute intricate task planning and reasoning. \cite{gong2023mindagent} investigates the application of LLMs in scenarios involving multi-agent coordination, covering a range of diverse task objectives. \cite{zeng2022socratic} presents a modular framework that employs structured dialogue through prompts among multiple large pretrained models. Moreover, specialized LLMs for particular applications have been explored, such as HuggingGPT \cite{shen2023hugginggpt} for vision perception tasks, VisualChatGPT \cite{wu2023visual} for multi-modality understanding, Voyager \cite{wang2023voyager} and \cite{zhu2023ghost}, SheetCopilot \cite{li2023sheetcopilot} for office software, and Codex \cite{Codex} for Python code generation.
Furthermore,AutoGPT\cite{AutoGPT} demonstrates the ability to autonomously complete tasks by enhancing AI models, but it is a single-agent system and does not support multi-agent collaboration. In contrast, BabyAGI\cite{babyagi} uses multiple agents to manage and complete tasks, with each agent responsible for different task modules such as creating new tasks, prioritizing the task list, and completing tasks. Multi-agent debate research includes works\cite{Liang_Encouraging_2023}\cite{Du_Improving} indicating that debates among multiple LLM instances can improve reasoning and factuality, but these methods typically lack the flexibility of tool and human involvement. AutoGen\cite{Wu_Autogen_2023}, as a general infrastructure, supports dynamic dialogue modes and a broader range of applications, demonstrating potential in advancing this field.

\noindent\textbf{3D Urban Scene Generation.} Scene-level content generation presents a challenging task, unlike the impressive 2D generative models primarily targeting single categories or common objects, due to the high diversity of scenes. Semantic image synthesis, as exemplified by \cite{Esser_Taming_2021,hao2021GANCraft,Mallya_World-Consistent_2020,Park_Semantic_2019}, has shown promising results in generating scene-level content in the wild by conditioning on pixel-wise dense correspondence, such as semantic segmentation maps or depth maps. Recent works such as \cite{chai2023persistent,SceneDreamer,InfiniCity} have realized infinite-scale 3D consistent scenes through unbounded layout extrapolation. Additionally, in-depth research has been conducted on using procedural content generation (PCG) techniques to generate natural scenes \cite{ecosystems2022,rivers} and urban scenes \cite{lipp2011interactive,talton2011metropolis,Parcels,UrbanPattern2013}. For example, PMC \cite{PMC} proposed a procedural method based on 2D ocean or city boundaries to generate cities. It employs mathematical algorithms to generate blocks and streets and utilizes subsequent techniques to generate the geometric shapes of buildings. While traditional computer graphics methods can generate high-quality 3D data, all parameters must be predefined during the procedural generation process. Since the generated 3D data is subject to rule limitations and exhibits a certain degree of deviation from the real world, this significantly constrains its flexibility and practical utility.

\section{Methods}
\label{headings}
The overall objective is to generate high-quality, controllable, and editable interactive large-scale city scenes in multiple stages based on natural language instructions and multimodal inputs. To achieve this goal, we propose the \mymethod{} framework, where we use procedural generation(PCG) to construct the entire scene and leverage our designed multi-agent system to enable the LLM to parse parameters and act as an agent based on natural language instructions. In Sec \ref{sec:generate}, we described the relevant PCG modules we have collected and created based on the requirements of the urban scene generation task. In Sec \ref{sec:procotol}, we describe the PCG management protocol we designed to standardize and manage the PCG modules, ensuring their compatibility and scalability. In Sec \ref{sec:agent}, we present our multi-agent system, optimized with visual feedback, which supports the flexible, fast, and accurate construction of urban scenes based on language descriptions and multimodal inputs. In Sec \ref{sec:dataset}, we showcase the diverse, high-quality datasets generated through this process, including various formats.

\subsection{Key PCG Modules in Urban Scene Generation}
\label{sec:generate}

To enhance the realism of the generated urban scenes, we categorize the scene elements into roads, buildings, greenery, pedestrians, traffic flow, and water bodies. Among these, roads, buildings, pedestrians, and traffic flow are of particular importance, warranting dedicated PCG modules for each. To support these modules, we have meticulously curated and, in some cases, crafted specialized PCGs for comprehensive integration into the urban environment.

\begin{figure}[!tbp]
\centering
\includegraphics[width=1\linewidth]{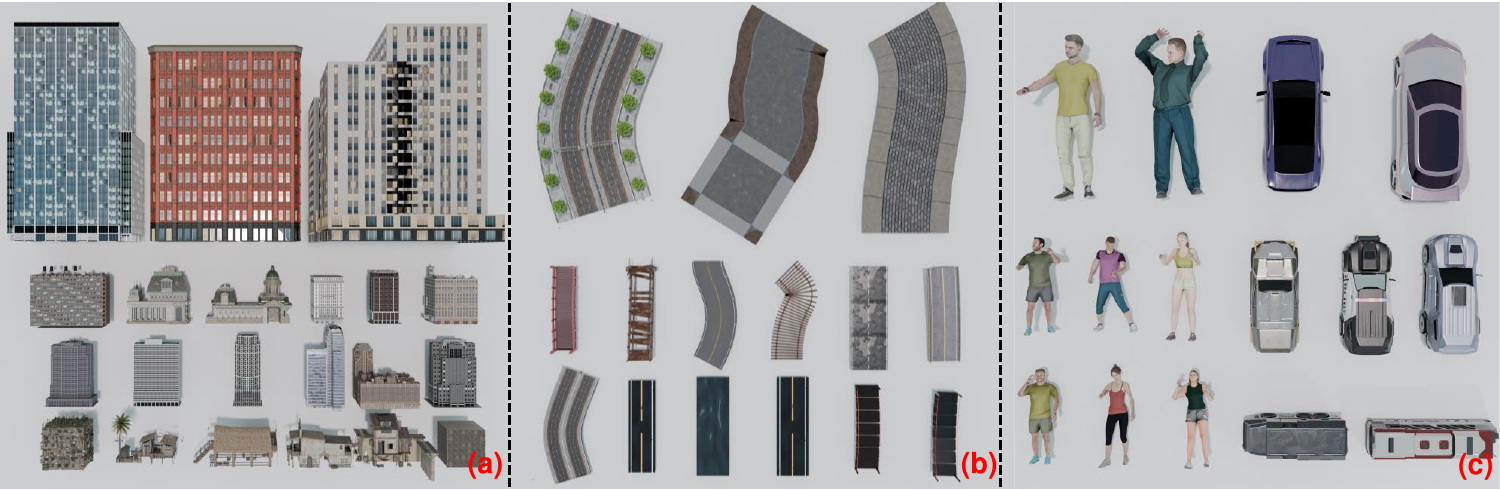}
\caption{Presentation of PCG modules involved in urban scene generation: building generation PCG effects (a), road generation spline-based PCG effects (b), and dynamic pedestrian and traffic flow generation PCG effects (c).}
\label{fig:pcg}
\end{figure}

\noindent\textbf{Building Generators.} We offer various PCG methods for building generation, such as using L-systems\cite{Prusinkiewicz_Lindenmayer_1990} to simulate the basic structure of buildings, akin to the growth process of plants. Through recursive algorithms starting from simple rules, complex architectural forms are progressively generated. Noise functions, such as Perlin noise\cite{Perlin_1985}, are employed to introduce randomness and variation, preventing the buildings from appearing overly regular or artificial. Fig. \ref{fig:pcg}.(a) illustrates a series of buildings generated using these PCGs. In addition, we have curated and annotated high-quality, artistically crafted building models, recording information such as the building's footprint and style. For retrieval, we utilize a pre-trained CLIP\cite{Radford_2021} model, integrating the matching models into the urban scene. Due to space constraints, further details can be found in the appendix.

\noindent\textbf{Roads Generators.} We provide various road generators (Fig. \ref{fig:pcg}.(b)) capable of creating common types of roads such as asphalt roads, highways, and secondary roads. These generators support control over parameters such as road width, structure, and style. Roads can be generated based on curves, and some PCGs also generate road materials based on the road's mesh.

\noindent\textbf{Pedestrian \& Vehicle Flow Generators.}
For pedestrian and traffic flow generation, we also provide relevant PCG modules (Fig. \ref{fig:pcg}.(c)), allowing for additions based on path planning and coordinate inputs. Detailed interactions and input formats are available in the appendix.

\subsection{PCG Management Protocol}
\label{sec:procotol}

Recently, existing works 3D-GPT \cite{3D-GPT}, Scenecraft \cite{hu2024scenecraft} and SceneX \cite{SceneX} have demonstrated the potential of using large language models (LLMs) to orchestrate PCG programs for general scene generation. While these methods have shown promising performance, the quality of the generated scenes often depends heavily on the quality and diversity of the collected PCG assets. However, integrating diverse plugins from various sources is a significant challenge due to the heterogeneity in their interfaces, formats, and functional capabilities. In Sec \ref{sec:generate}, we encountered issues with data format incompatibility among the PCGs we collected.Therefore, we propose a universal PCG management protocol that seamlessly integrates infinite plugin assets to break down the asset barriers in LLM-driven procedural scene generation. The proposed method includes three aspects: \textit{Dynamic API Conversion Interface} and \textit{Structured Encapsulation}.

\noindent\textbf{Dynamic API Conversion Interface.} Like HuggingGPT 
\label{sec:api}
\cite{shen2023hugginggpt} that integrates external algorithm API with LLMs, the motivation of our Dynamic API Conversion Interface is to provide a universal conversion method that transforms PCG plugins into a format callable by LLM agents. It can empower the LLMs to orchestrate and execute procedural scene generation tasks across a wide range of assets. During the PCG integration process, when a format conversion is needed, the corresponding interface can be automatically invoked for the conversion. For instance, we can convert the curve of the road in this step to a plane, allowing the downstream PCG to continue the subsequent tasks. This dynamic API conversion interface serves as a bridge between different APIs, providing communication interfaces for various API formats. By dynamically adjusting these interfaces, we can achieve the flexible combination of different PCGs.We provide detailed conversion examples and an analysis of interface format types in the appendix.

\noindent\textbf{Structured Encapsulation.} Since LLM cannot directly use PCG through Blender, it is necessary to encapsulate PCG into action functions that are executable by LLM. Additionally, we need to parameterize each PCG so that LLM can plan tasks more accurately based on the functional boundaries of the PCG. Moreover, encapsulating PCG into action functions presents significant technical barriers for beginners, particularly in terms of coding knowledge and 3D modeling expertise. To help beginners quickly and easily create their own action functions,we propose a method of structured encapsulation. For each PCG, the encapsulation, denoted as Encapsulation $S$, follows a similar structure defined as $S_{\text{i}}$($C_{\text{i}}$, $D_{\text{i}}$, $I_{\text{i}}$, $L_{\text{i}}$, $R_{\text{i}}$), where $i$ distinguishes the $i$-th action function. The components of the protocol are defined as follows:
\begin{itemize}
    \item[] $\bullet$ \textbf{Classname} The name of the action function, used for indexing the action function.
    \item[] $\bullet$ \textbf{Description} A detailed objective description of an action function.
    \item[] $\bullet$ \textbf{Input} A detailed objective description of the input to the action function.
    \item[] $\bullet$ \textbf{Limitation} An objective description of the functional constraints of the action function.
    \item[] $\bullet$ \textbf{Run} Function name used to execute the action function and return the result.
\end{itemize}
We provide users with a straightforward encapsulation document. By acquainting themselves with the process of extracting API information from Blender, users can seamlessly follow the document's instructions to encapsulate their own functional actions.

\begin{figure*}[!tbp]
\centering
\includegraphics[width=1\linewidth]{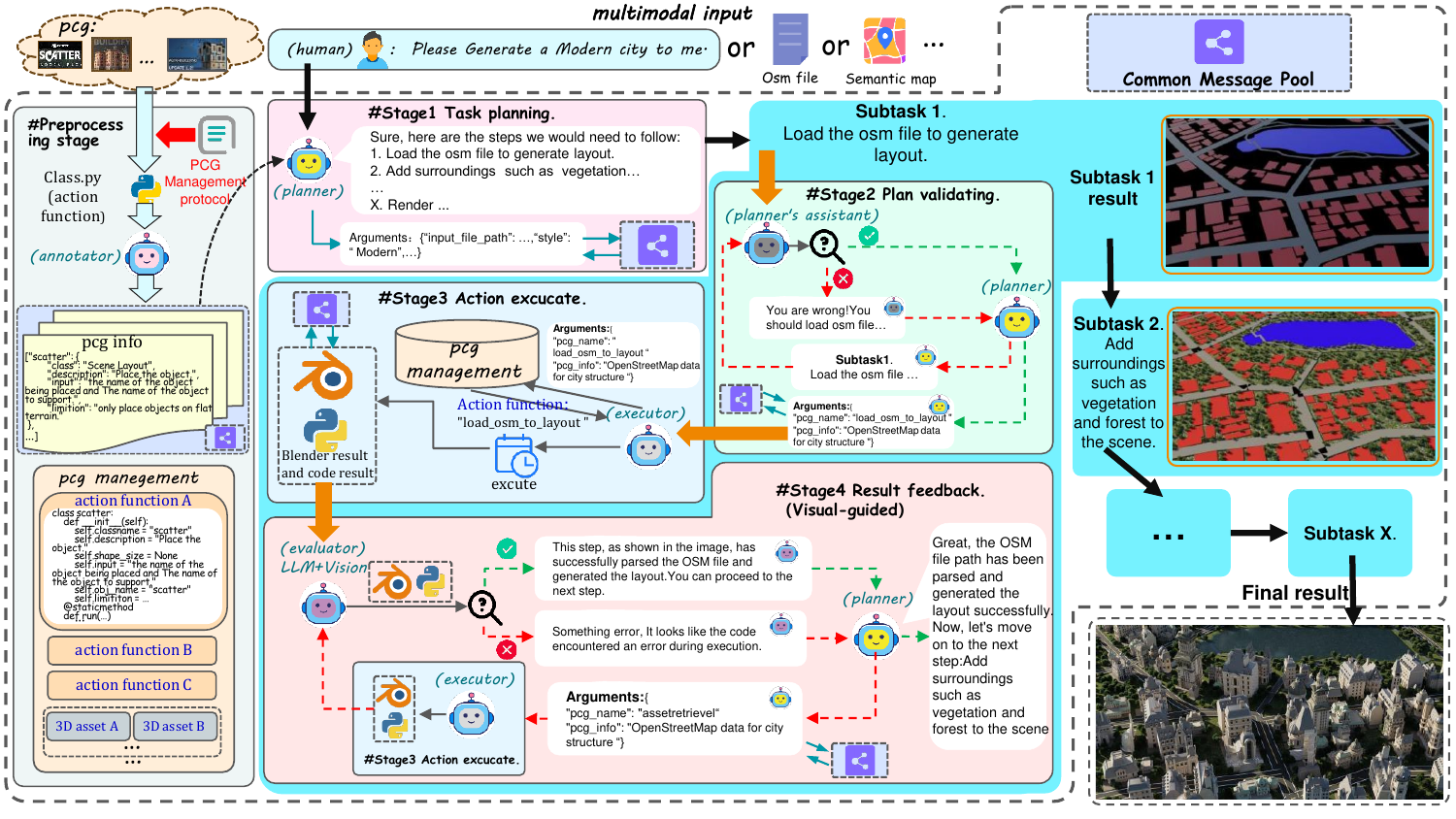}
\vspace{-0.5cm}
\caption{Multi-agent Workflow: Detailed demonstration of collaboration and communication across various stages. }
\vspace{-0.5cm}
\label{fig:pipeline}
\end{figure*} 
\subsection{Multi-Agent Framework}
\label{sec:agent}
Fig. \ref{fig:pipeline} illustrates the overall generation workflow of \mymethod{}, with its core components comprising the PCG management protocol module and the multi-agent module. Due to the need for efficient multi-stage interactive generation, such as the blueprint for layout planning, the geometry node for asset scatter, and numerous parameters for asset placement, color, size, and height, this renders a naive LLM framework inadequate for handling Blender's complex and diverse action. This framework mainly consists of four agents: annotator, planner, executor, and evaluator. The annotator labels all actions with multiple tags and stores the annotated action information in the common message pool. The planner formulates the overall task pipeline using user-provided textual information and determines the action required for different sub-tasks. The executor manages all action functions and uses the annotated action functions to process sub-tasks of PCG or asset manipulation in Blender. The evaluator assesses the correctness of the current sub-task by obtaining scene information from Blender, such as object location and rendered images.

\noindent\textbf{Multimodal Input through Multi-Agent Framework.}
Unlike simple LLM frameworks, our Multi-Agent Framework can find feasible solutions through multiple rounds of interaction and visual feedback, meaning that with only a few plugins, the target task can be accomplished, avoiding the burden to reinvent the wheel. This also simplifies city generation through multimodal inputs. For example, when we import a semantic map into Blender, it is stored as a point cloud containing semantic information, but we only have face-based building generation actions. At this point, the Multi-Agent Framework will find a tool to convert vertices into faces, complete the conversion, and then use face-based building generation actions to accomplish the task.

\noindent\textbf{Annotator Agent.}
For large-scale city generation tasks, a substantial number of action functions are required. Therefore, effectively managing these action functions is crucial. To ensure each agent can efficiently access all action of different hierarchies, we use the Annotator to label them.More details of the annotations are provided in the appendix, along with relevant examples.

\begin{figure*}[!tbp]
\centering
\includegraphics[width=0.98\linewidth]{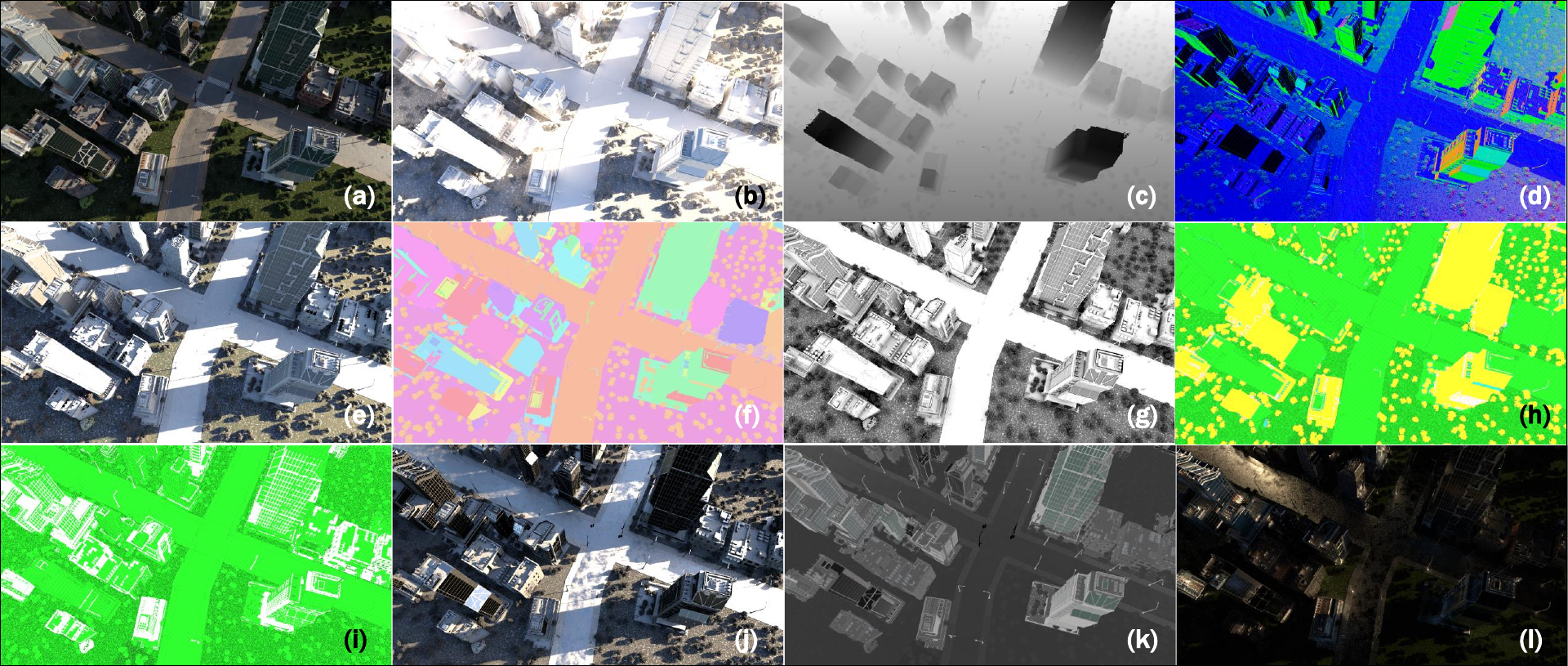}
\caption{For each image (a), we have a high-resolution mesh (b), which readily yields Depth (c), Surface Normals (d), Diffuse Map (e), Instance Segmentation masks (f), Ambient Occlusion (g), Cryptomatte Object/Material Mask (h/i), and Glossy Direct/Color/Indirect (j/k/l).We used a 1920 × 1080 resolution with 10,000 random samples per pixel, a standard setting in Blender that effectively eliminates sampling noise, resulting in a high-quality final image.}
\vspace{-0.5cm}
\label{fig:dataset}
\end{figure*} 
\noindent\textbf{Planner Agent.}
We consider PCG-based city scene generation as an open-loop planning task with flexible steps. Specifically: (i) Termination of the city scene generation task depends on meeting the user’s requirements; (ii) City scene generation tasks are a series of sequentially arranged action functions, where the order of these actions significantly impacts the final outcome. To address these challenges, we propose a dynamic planner. At the beginning of the task, the planner formulates a rough workflow as a reference. During the execution of specific sub-tasks, the planner plans the next action based on the current sub-task goals and the workflow, until the user’s requirements are met.

When the planner directly receives user input, it needs to translate the user's intent into a series of referable executable actions with additional explanations, which will be stored in the common message pool. To achieve this, we prompt the planner to produce a preliminary action plan, serving as the workflow. Specifically, we utilize the labeled information \( L \) from the common message pool, the user input \( I \), and the planner's guidance document \( D \) as inputs, allowing the planner to generate an ordered workflow \( W \). This process is formalized as follows:
\begin{equation}
W \leftarrow \text{Planner}(L, I, D)
\end{equation}
During the execution of the \( t \)-th sub-task, the planner needs to formulate actions required for the \( (t+1) \)-th sub-task. To ensure the accuracy and coherence of actions, the planner refers to the workflow. Specifically, based on the ordered workflow \( W \), sub-task input \( I \), the labeled information \( L \) from the common message pool, and the planner's guidance document \( D \), the planner infers the next action \( A_{t+1} \), where \( A_{t+1} \) stands for the action of the \( (t+1)\)-th sub-task.
\begin{equation}
A_{t+1} \leftarrow \text{Planner}(L, I, D, W).
\end{equation}

\noindent\textbf{Executor Agent.}
To achieve Interactive Workflow in Blender, we deploy the Executor within the Blender environment. All agents send action execution commands to the Executor through a local backend server. As mentioned in Sec \ref{sec:generate}, we transform PCG plug-ins into executable action functions using structured encapsulation. This allows the Executor to initialize all actions flexibly. To enable agents to precisely control actions, each action function is structurally recorded in JSON format. For example, the scale\_object action function is documented as follows:
\lstset{basicstyle=\footnotesize, breaklines=true}
\begin{lstlisting}
scale_object_doc = {name: scale_object, description: Scale an object,
    parameters:{scale_factor:{type:tuple, description:scale factor},
        scaled_obj_name:{type:str, description:scaled object name}}}
\end{lstlisting}

During the execution of the \( t \)-th sub-task, the Executor uses the action document \(D\) and the sub-task input \(I\) to generate the action arguments \(Arguments\). The action \(A_t\) is then executed in Blender based on the arguments. Here, \( A_{t+1} \) represents the action for the \((t+1)\)-th sub-task, \( S_{t+1} \) represents the Blender state for the \((t+1)\)-th sub-task, and \( S_t \) represents the Blender state for the \( t \)-th sub-task.
\begin{multline}  
\text{Arguments} \leftarrow \text{Executor}(I, D), \\
S_{t+1} \leftarrow \text{Blender}(S_t, A_t, \text{Arguments})
\end{multline}

\noindent\textbf{Evaluator Agent.}
To address the limitations of textual feedback in urban scene generation tasks, we designed an Evaluator with visual feedback based on GPT-4V\cite{openai-gpt4vision}. Specifically, we first render the generated scene as an image. Then, we provide both the image and the current sub-task text input to the Evaluator, guiding it with prompts to assess whether the scene's geometry and materials match the sub-task expectations. If the Evaluator determines that the rendered image is consistent with the sub-task text input in terms of geometry and materials, the evaluation ends. However, if the Evaluator identifies errors, it can pass this information to the Planner for improvements. This process is formalized as follows:
\begin{equation}
\text{R} \leftarrow  \text{Evaluator}(\text{GPT-4V}(img, I, D)),
\end{equation}
where \( R \) stands for the Evaluator result, \( img \) stands for the rendered image, \( I \) stands for the sub-task input, and \( D \) stands for the Evaluator's guidance document.

\subsection{Urban Scene Dataset Synthetic}
\label{sec:dataset}
Based on the scenes we have built, we can generate an unlimited number of high-quality city scene datasets, which are rendered using Blender’s physically-based path tracing renderer, Cycles, as shown in Fig. \ref{fig:dataset}. The scenes that can be generated are infinite, and thus the dataset is also unlimited. Previous related datasets often faced issues such as high production costs or lower quality. However, our implementation is highly efficient and low-cost. We compare the advantages of our dataset with other urban scene datasets in the appendix and present more details about the datasets. We anticipate that users will be able to easily extend our codebase to generate various city scene datasets, meeting the needs of applications in related fields.

\section{Experiments}
\label{others}

The goals of our experiments are threefold: (i) to verify the capability of \mymethod{} for generating highly realistic large-scale city with different modes of input; (ii) to prove that the Muti-Agent framework and pcg protocol we designed are effective; (iii) to compare different LLMs on the proposed benchmark.
\subsection{Benchmark Protocol}
\noindent\textbf{Dataset.} To evaluate the effectiveness of the proposed \mymethod{}, we collect 50 city Semantic Maps, 50 city Height Fields, and 50 city OSM files in XML format. We also collect 50 sets of descriptions about city styles and weather. Then, we feed them to our \mymethod{} to generate corresponding city models, which are used to perform quantitative and qualitative comparisons.

\noindent\textbf{Models.} When generating and editing the 3D scenes, we adopt the leading GPT-4 as the large language model with its public API keys. To ensure stable output from the LLM, we set both the decoding temperature and the seed to 0.

\begin{figure}[!tbp]
\centering
\includegraphics[width=1\linewidth]{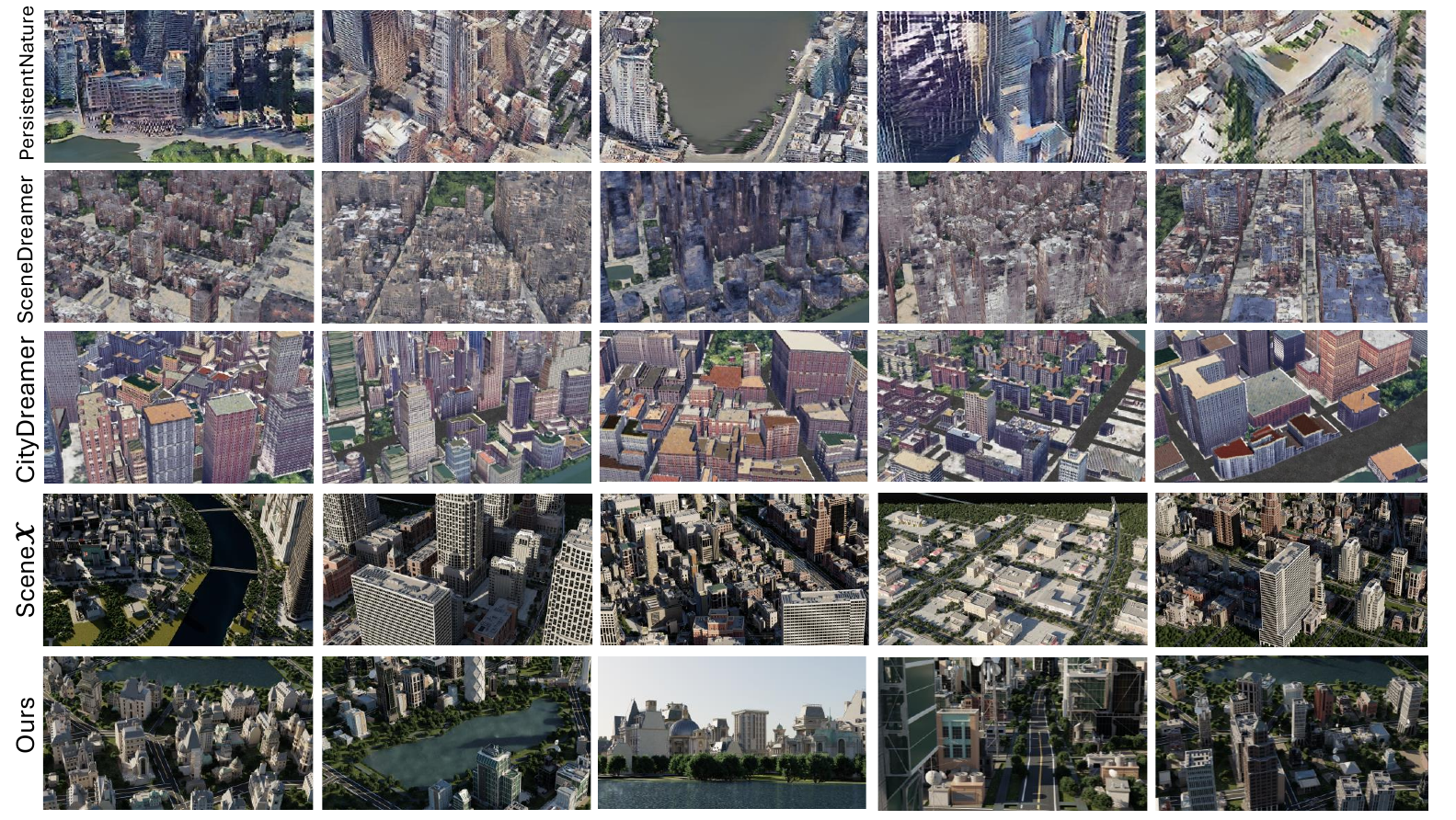}
\caption{Comparative results on city generation. Issues with unreasonable geometry are observed in previous works, while our method performs well in generating realistic large-scale city scenes.}
\vspace{-0.5cm}
\label{fig:city_compare}
\end{figure} 

\begin{figure}[!tbp]
\centering
\includegraphics[width=1\linewidth]{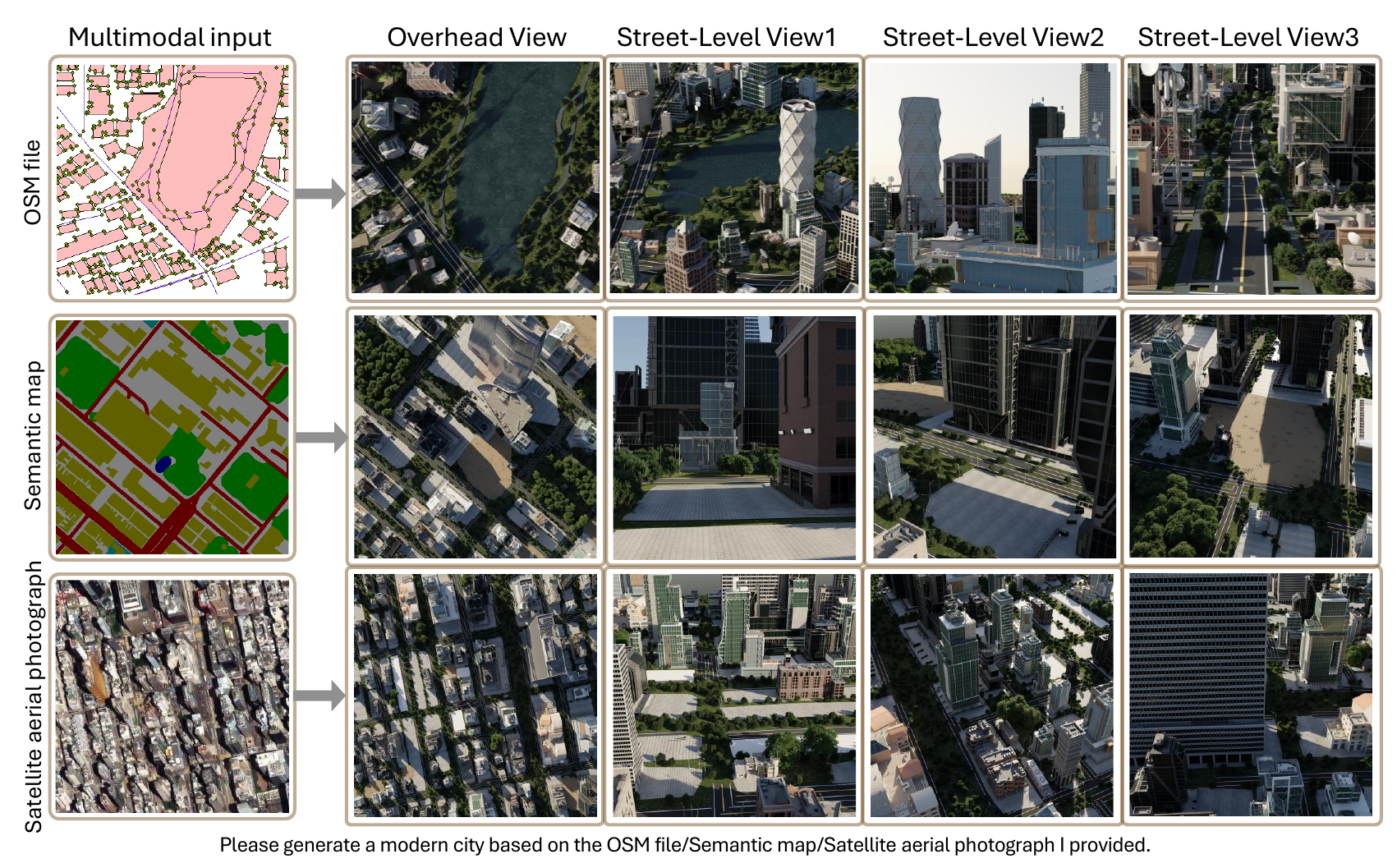}
\caption{Urban scene generation with multimodal inputs, where we present an overhead view aligned with the multimodal input perspective, along with three street-level views.}
\vspace{-0.5cm}
\label{fig:multimodel}
\end{figure} 

\begin{table}[!tbp]
\caption{Comparing the performance of different language models in city generation. The aesthetic score represents the intuitive aesthetic rating, and the rationality score is the logical coherence rating.Where "AS" represents the average score (from 30 volunteers), and "AES" represents the average expert score (from 5 graphics experts).}
\label{tab:ASAES}
\centering
\resizebox{\columnwidth}{!}{
\begin{tabular}{lcccccl}
\hline
\multicolumn{1}{c}{Model} & \multicolumn{2}{c}{AS} &           & \multicolumn{2}{c}{AES} &  \\ \cline{2-3} \cline{5-6}
                       & aesthetic score              &  rationality score             &           & aesthetic score              &  rationality score             &  \\ \hline
CityDream\cite{xie2023citydreamer}        &             2.80  &           3.05   &           &             2.65 &           3.40  &  \\
PersistentNature\cite{chai2023persistent}           &             1.30 &            1.40 &           &            1.55  &           1.35  &  \\
SceneDreamer\cite{SceneDreamer}            &             1.30 &           1.63  &           &            1.30  &           1.35  &  \\
Scene$\mathcal{X}$\cite{SceneX}           &            3.73  &            3.63 &           &      3.50  &          3.45   &  \\
\textbf{\mymethod{}(Ours)}             & \textbf{4.30}            & \textbf{4.35}          &           & \textbf{4.15}             & \textbf{4.30}     \\ \hline
\end{tabular}
\vspace{-0.5cm}
}
\end{table}

\noindent\textbf{Metrics:} We use Executability Rate (ER@1) and Success Rate (SR@1) to evaluate the capabilities of LLMs on our \mymethod{}. The former measures the proportion of proposed actions that can be executed, and the latter is used to evaluate action correctness \cite{chen2021evaluating}. Additionally, 
we use a unified evaluation standard as a reference. We categorize the aesthetics of city scenes into five levels: Poor (1 points)/Below Average (2 points)/Average (3 points)/Good (4 points)/Excellent (5 points).

\subsection{Main Result}
\noindent\textbf{Urban Scene Generation with Multimodal Inputs.} 
We first show the ability of \mymethod{} to generate large-scale urban scenes, as depicted in Fig. \ref{fig:multimodel}. The results show that \mymethod{} is capable of generating highly realistic urban scenes using multimodal data inputs, including OSM data, semantic maps, and satellite images, demonstrating its effectiveness and flexibility in urban scene generation.

We also compare our method with other city generation approaches, as shown in Fig. \ref{fig:city_compare}. The results indicate that PersistantNature\cite{chai2023persistent} and InfiniCity\cite{InfiniCity} have severe deformation issues throughout the entire scene. While SceneDreamer\cite{SceneDreamer} and CityDreamer\cite{xie2023citydreamer} demonstrate improved structural consistency, their building quality remains relatively low. While Scene$\mathcal{X}$\cite{SceneX} achieves high quality, it encounters issues with overlapping assets and a high duplication rate of buildings. In contrast, the city generated by \mymethod{} demonstrates a regular geometric structure and high quality, which is devoid of overlapping buildings and exhibits minimal repetition.

\noindent\textbf{Aesthetic Evaluation.} 
To better assess the quality of cities generated by \mymethod{}, we collect results from various related works on urban generation and invite 30 volunteers and 5 experts in 3D modeling to evaluate these works aesthetically. To ensure fairness, we anonymize all results. As shown in Tab. \ref{tab:ASAES}, the cities generated by \mymethod{} attain a "Good" level in aesthetic scoring, a distinction not achieved by other works, demonstrating its highly realistic capabilities in city generation.

\begin{figure}[!tbp]
\centering
\includegraphics[width=1\linewidth]{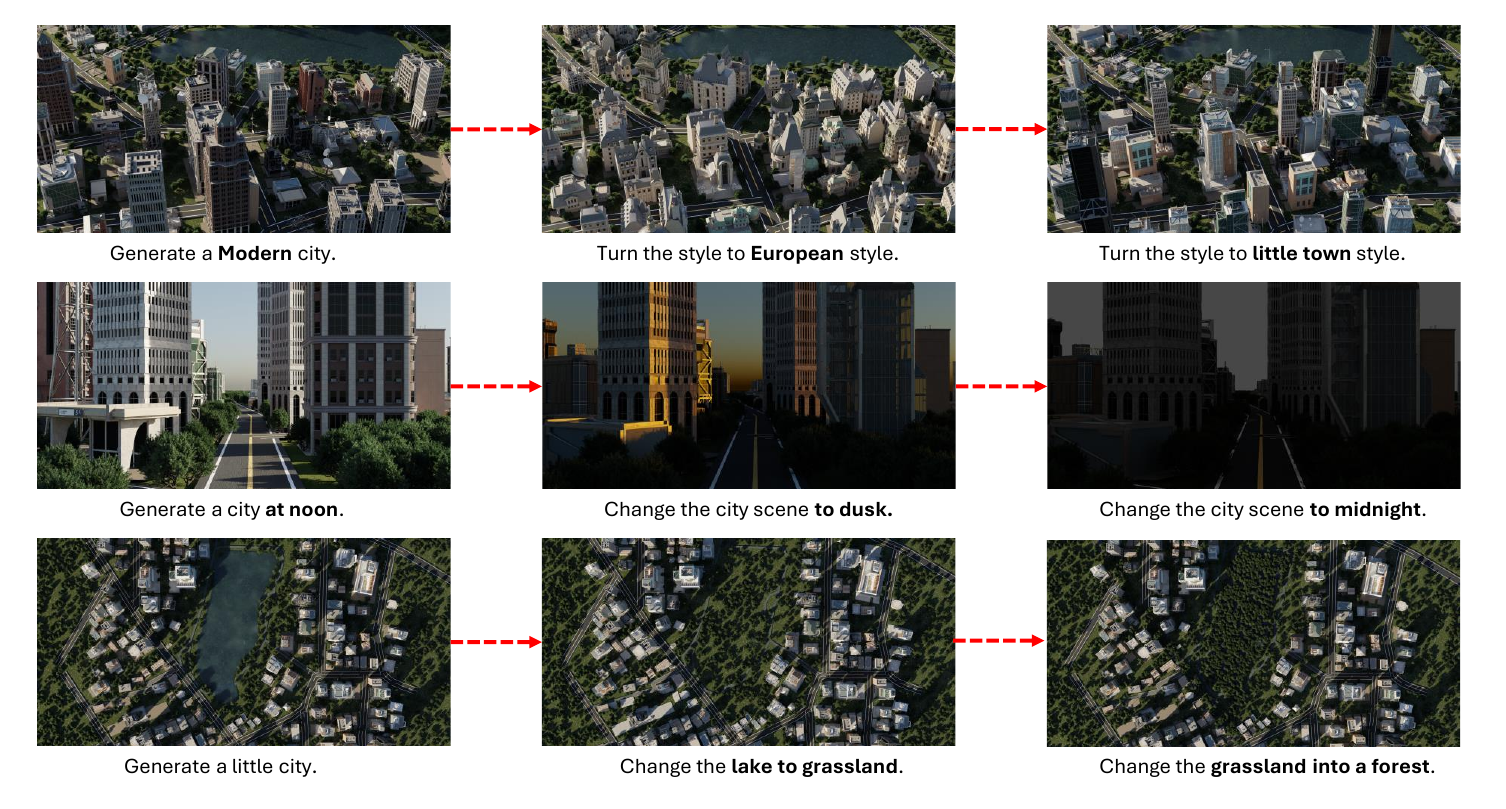}
\caption{Performance of \mymethod{} in scene customization and refinement editing}
\vspace{-0.5cm}
\label{fig:edit}
\end{figure}

\noindent\textbf{Specific Refinement Editing.} 
\mymethod{} supports specific refinement editing for scene customization, involving asset manipulation, weather adjustment, and style modification. We conduct relevant experiments, as depicted in Fig. \ref{fig:edit}. Based on the results, it's clear that \mymethod{} performs well in accurately controlling urban scenes to meet input requirements consistently.

\begin{table}[ht]
\centering
\caption{Ablation Study Results for Different Components of the Protocol. \label{tab:Ablation studies}}
\setlength{\tabcolsep}{7pt}
\begin{tabular}{ccc|cc}
\hline
 Description.        & Input        & Limitation  &  ER@1  &  SR@1 \\ \hline
             &              & \checkmark  & 36.00  & 41.67 \\
             & \checkmark   & \checkmark  & 37.00  & 51.35\\
 \checkmark  &              &             & 44.00  &  56.82 \\  
 \checkmark  & \checkmark   &             & 69.00  & 60.87\\
 \checkmark  &              & \checkmark  & 73.00 & 61.64\\
 \checkmark  & \checkmark   & \checkmark  &  \textbf{94.00}  & \textbf{82.98}\\ \hline
\end{tabular}
\end{table}

\begin{table}[ht]
\centering
\caption{Comparing the performance of different language models in city generation.\label{tab:ERSR}}
\setlength{\tabcolsep}{14pt}
\begin{tabular}{lcccl}
\hline
{Model}  & ER@1              & SR@1       \\ \hline
Llama2-7B\cite{touvron2023llama}          & 27.00             & 59.26            &  \\
Mistral\cite{jiang2023mistral}            & 78.00             & 61.54               &  \\
Gemma-2B\cite{gemmateam}           & 9.00             & 33.33                    &  \\
Gemma-7B\cite{gemmateam}            & 39.00             & 69.23                    &  \\
GPT-3.5-turbo\cite{brown2020language}      & 72.00             & 75.00                   &  \\
GPT-4\cite{openai2023gpt4}              & \textbf{91.00}            & \textbf{81.32}               \\ \hline
\end{tabular}
\end{table}

\subsection{Ablation Study of PCG Protocol.}
\noindent\textbf{Analysis of Different Components.} To assess the impact of each component of the structured encapsulation on the overall system, we conduct ablation experiments on individual parts of the structured encapsulation, as shown in Tab. \ref{tab:Ablation studies}. The table demonstrates that when encapsulating PCG, adding Description, Input, and Limitation all boost ER@1 and SR@1. Notably, including Description leads to the highest increase, with SR@1 rising by 57.00\% and ER@1 by 31.63\%, significantly enhancing the system's Executability Rate and Success Rate. After adding Input, SR@1 and ER@1 increase by up to 21.34\% and 25.00\%, respectively. Similarly, with the inclusion of Limitation, SR@1 and ER@1 see maximum increases of 22.11\% and 25.00\%. This suggests that incorporating Input and Limitation can improve the system's Executability Rate and Success Rate.

\noindent\textbf{Comparing Agent Frameworks with Different LLMs}
To evaluate the effects of different large language model variants on multi-agent frameworks, we assessed the system's ER@1 and SR@1 using various LLM versions. As not all open-source models have visual perception capabilities, we used GPT-4-Vision-Preview uniformly for visual feedback. To maintain experimental stability, the temperature and seed for all LLMs were set to 0. The experiment utilized 50 description sets from Section 4.1 dataset. Results are shown in Tab. \ref{tab:ERSR}. GPT-4 achieves the highest scores in both ER@1 and SR@1, with scores of 91.00\% and 89.01\%, respectively. Mistral ranks second in ER@1, with a score of 78.00\%, while GPT-3.5-turbo and Gemma-7B rank second and third in SR@1, with scores of 75.00\% and 69.23\%, respectively. It is evident that Mistral and Gemma-7B, two open-source large language models, perform comparably to GPT-3.5-turbo but still fall short of GPT-4's performance.

\section{Conclusion}
\label{Conclusion}
This paper presents a streamlined and productive pipeline for controllable 3D urban scene generation.
Rather than creating scenes from scratch using neural generative models, we resort to the procedural generation paradigm. 
We design a management protocol to enable flexible integration and free calling.
A multi-agent framework powered by LLMs is introduced to automate the creation process according to multi-modal instructions, such as OSM. Our generated scenes are interactable and superior in controllability, diversity, and fidelity, making them well-suited as base environments for simulators. Our method has the potential to significantly advance the field of embodied intelligence.

\setcounter{section}{0}
\renewcommand\thesection{\Alph{section}}
\renewcommand\thesubsection{\thesection.\arabic{subsection}}

\clearpage
\twocolumn[{%
\centering
}]

\section*{\relsize{+1}\bfseries Appendix}
\addcontentsline{toc}{chapter}{Appendix} 

\vspace{1em} 


\section{Implementation Details}
\label{sec:details}

\subsection{Key PCG Modules}
\label{sec:pcg}

We roughly divide the city scene generation into the following stages:
\begin{equation}
S = \{ S_{\text{layout}}, S_{\text{terrain}}, S_{\text{building}}, S_{\text{road}}, S_{\text{nature}}, S_{\text{background}} \},
\end{equation}
where \(S_{\text{layout}}\) represents layout generation, \(S_{\text{terrain}}\) represents terrain generation, \(S_{\text{building}}\) represents building generation, \(S_{\text{road}}\) represents road generation, \(S_{\text{nature}}\) represents nature elements generation (including vegetation, rivers, lakes, etc.), and \(S_{\text{background}}\) represents background generation (e.g., weather and sunlight).

The dependency relationships between stages can be expressed as a directed graph:
\begin{equation}
G = (V, E),
\end{equation}
where \(V = S\) is the set of stages, and \(E\) is the set of dependencies. For example, \( (S_{\text{layout}}, S_{\text{building}}) \in E \) indicates that building generation depends on layout generation. However, certain stages, such as \(S_{\text{building}}\) and \(S_{\text{road}}\), do not require a specific order.

Each stage \(S_i\) comprises a set of PCG (Procedural Content Generation) modules:
\begin{equation}
S_i = \{ M_{i, 1}, M_{i, 2}, \dots, M_{i, k_i} \},
\end{equation}
where \(M_{i, j}\) is the \(j\)-th module in the \(i\)-th stage, and \(k_i\) represents the total number of modules in that stage.

Given specific requirements \(R\), the system selects PCG modules that satisfy the requirements from each stage:
\begin{equation}
M_{\text{selected}} = \bigcup_{i \in \{1, 2, \dots, n\}} \{M_{i, j} \mid M_{i, j} \text{ satisfies } R \}.
\end{equation}

The generation process for different cases can be represented as specific sequences of selected modules:
\begin{equation}
F_{\text{case}} = [M_{i_1, j_1}, M_{i_2, j_2}, \dots, M_{i_m, j_m}],
\end{equation}
where \(F_{\text{case}}\) denotes the generation pipeline for a particular case, and the sequence of modules is determined based on the requirements or task logic.

\begin{figure*}[!tbp]
\centering
\includegraphics[width=1\linewidth]{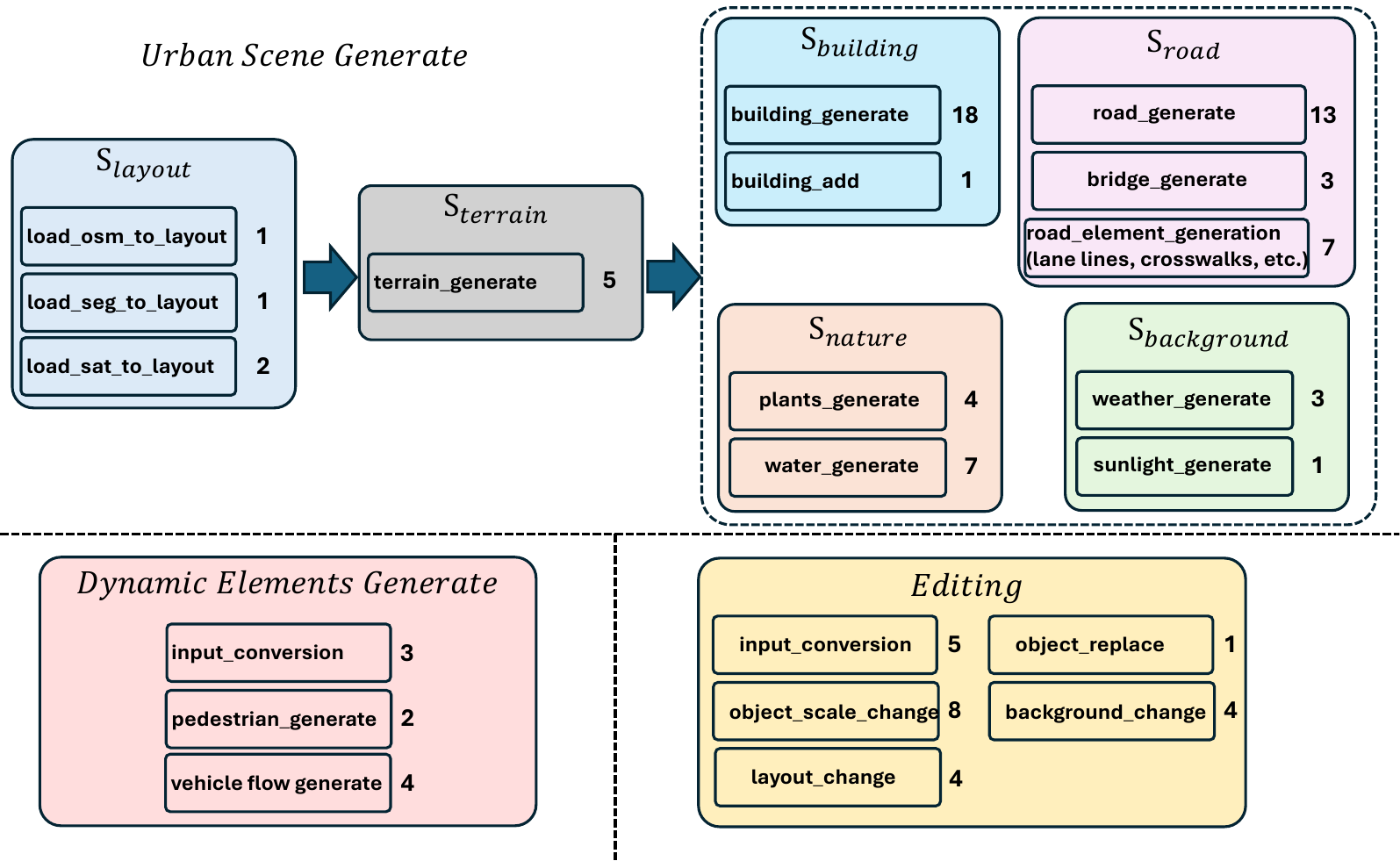}
\caption{Task phases and PCG module counts in the overall city generation and editing process, with multi-agent system guidance for task selection and execution.\\  \\ }
\vspace{-0.5cm}
\label{fig:app_generate}
\end{figure*}

\begin{figure*}[!tbp]
\centering
\includegraphics[width=1\linewidth]{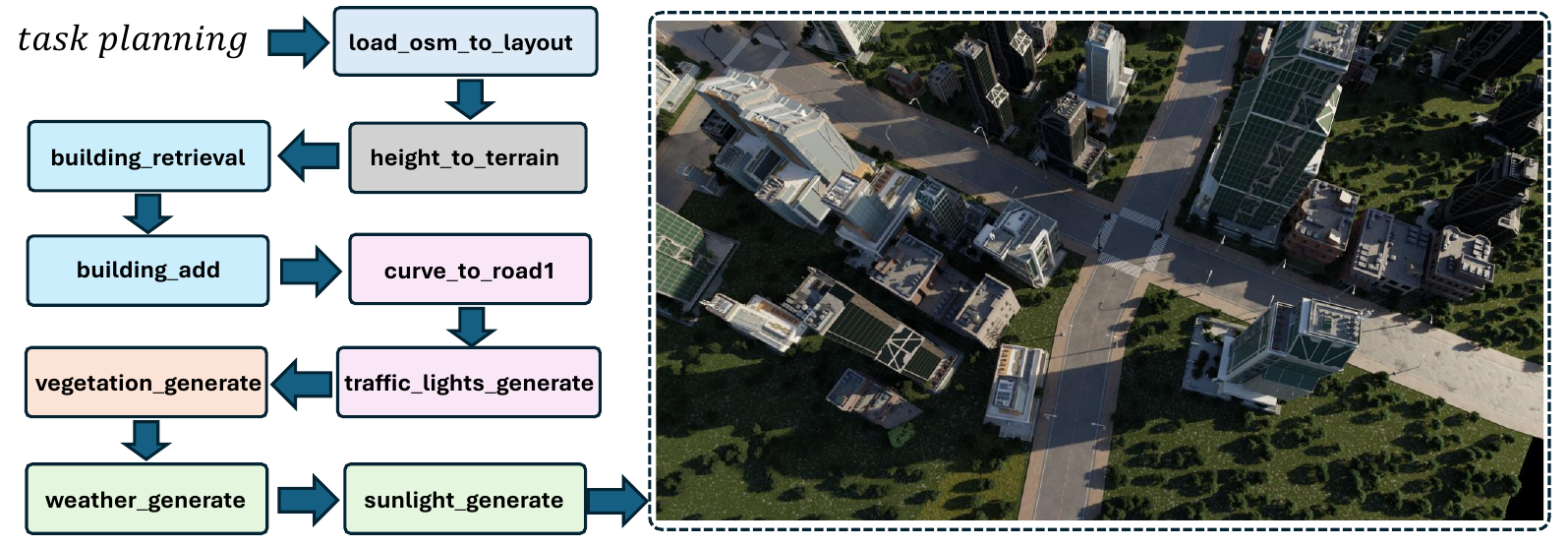}
\caption{Task phases and PCG module counts in the overall city generation and editing process, with multi-agent system guidance for task selection and execution.}
\vspace{-0.5cm}
\label{fig:app_generate_demo}
\end{figure*}

Fig.~\ref{fig:app_generate} illustrates the task phases and PCG module counts in the overall city generation and editing process we collected. During execution, specific requirements guide the selection and execution of tasks by a multi-agent system. Fig.~\ref{fig:app_generate_demo} presents an example case, demonstrating the generation process and the selected PCG modules involved.

\subsection{Dynamic Element Generation and Interactive Editing}
\label{sec:dynamic}

After the successful generation of the urban scene, we provide various implementations for interactive editing (e.g., editing specific elements in the scene) and the generation of dynamic elements (e.g., pedestrians, vehicle traffic). The entire workflow can be divided into two main parts: interactive input and specific execution. Interactive input focuses primarily on the extraction and transformation of information, such as determining the location or path for generating dynamic elements. For example, when we want to add a running character at a specific location in the scene, we can input this information by describing the path location or directly drawing a curve in the scene. This information is then processed through the corresponding PCG module, generating the corresponding instance (e.g., a curve object) in the Blender scene. After converting the path into a curve in the scene, the additional parameters related to the running character (e.g., speed, posture, clothing) are used to generate and insert the "running character" into the scene using the corresponding PCG module. The following pseudocode Algorithm .\ref{alg:running_character} shows the specific implementation of adding a pedestrian. Similarly, the generation of vehicle traffic and other interactive editing processes also follow the pattern of first transforming input and then using the corresponding PCG module for execution. Fig. \ref{fig:app_generate} shows the number of PCGs involved in adding dynamic elements, as well as the number of PCGs supporting interactive editing functionalities. 

\begin{figure*}[!tbp]
\centering
\includegraphics[width=1\linewidth]{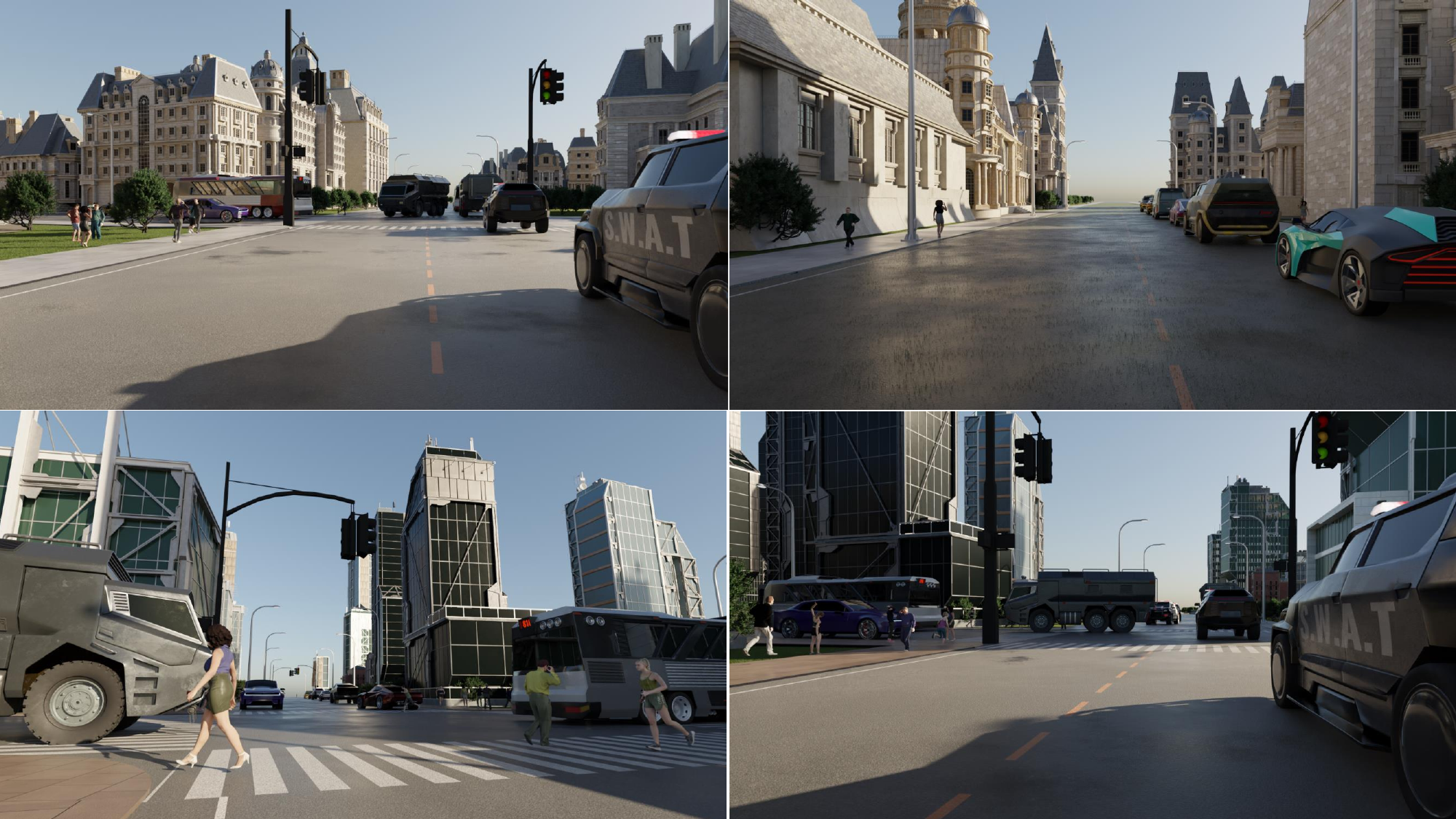}
\caption{Implementation results of adding dynamic elements.}
\vspace{-0.5cm}
\label{fig:renche_demo}
\end{figure*}

\begin{algorithm}
\caption{Generate Running Character in Scene}
\label{alg:running_character}
\begin{algorithmic}[1]
\STATE \textbf{Input:} Path data (\texttt{path = curve01}), dynamic element parameters (\texttt{crowd\_creation}, \texttt{walk\_run}, \texttt{style}, \texttt{category}, \texttt{model\_quality}, \texttt{random\_color})
\STATE \textbf{Output:} Generated running character placed in Blender scene

\STATE Let \texttt{path} = \texttt{curve01}.
\STATE Let \texttt{crowd\_creation} = ``single".
\STATE Let \texttt{walk\_run} = ``run".
\STATE Let \texttt{style} = ``casual".
\STATE Let \texttt{category} = ``run, male\_jog02".
\STATE Let \texttt{model\_quality} = ``high\_poly".
\STATE Let \texttt{random\_color} = \texttt{True}.

\STATE Create curve object \texttt{curve01} in Blender.
\STATE Set path of \texttt{curve01} as input path data.

\STATE \textbf{Parse parameters:}
\STATE \quad Assign \texttt{crowd\_creation} to ``single".
\STATE \quad Assign \texttt{walk\_run} to ``run".
\STATE \quad Assign \texttt{style} to ``casual".
\STATE \quad Assign \texttt{category} to ``run, male\_jog02".
\STATE \quad Assign \texttt{model\_quality} to ``high\_poly".
\STATE \quad Assign \texttt{random\_color} to \texttt{True}.

\FOR{each point in path (\texttt{curve01})}
    \STATE Calculate movement vector \(\mathbf{v}_{move}\) based on path curvature.
    \STATE Generate character model with specified parameters:
    \STATE \quad Apply \texttt{style} = ``casual".
    \STATE \quad Apply \texttt{category} = ``run, male\_jog02".
    \STATE \quad Apply \texttt{model\_quality} = ``high\_poly".
    \STATE \quad Assign color randomly if \texttt{random\_color} = \texttt{True}.
    \STATE Place character instance at position \(\mathbf{p}_{char}\) along path \texttt{curve01}.
    \STATE Set movement action: if \texttt{walk\_run} = ``run" then
    \STATE \quad Assign velocity \(\mathbf{v}_{run}\) for running.
    \STATE else
    \STATE \quad Assign velocity \(\mathbf{v}_{walk}\) for walking.
    \STATE end if
    \STATE Animate character along path based on velocity and movement vector.
\ENDFOR

\STATE \textbf{Output:} Place generated running character in scene at position \(\mathbf{p}_{char}\).
\STATE Return running character instance.
\end{algorithmic}
\end{algorithm}

\subsection{Details of PCG Management Protocol}
\label{sec:dynamic api}

\noindent\textbf{Dynamic API Conversion Interface.}
In Sec \ref{sec:api}, we discussed that the primary reason for the incompatibility among PCG modules lies in the difficulty of standardizing input and output data formats during the workflow transitions between different modules. Specifically, we first catalog all the possible input and output formats we have currently collected for the PCGs, as shown on the left side of Tab. \ref{apiformat}. These data formats represent the proportion of input and output formats in PCG. Based on this, we define a comprehensive and self-consistent dynamic API conversion interface, as shown on the right side of the Tab. \ref{apiformat}. During the PCG integration process, when a format conversion is needed, the corresponding interface can be automatically invoked for the conversion. For instance, we can convert the curve of the road in this step to a plane, allowing the downstream PCG to continue the subsequent tasks. This dynamic API conversion interface serves as a bridge between different APIs, providing communication interfaces for various API formats. By dynamically adjusting these interfaces, we can achieve the flexible combination of different PCGs.

\begin{table}[]
\caption{The proportion of API Input/Output Formats and Conversion Interface Statistics. A statistical overview of API input/output formats (in percentages) is presented on the left, where PIDF represents the proportion of input data format types and PODF represents the proportion of output data format types, while the conversion interface along with its descriptions is detailed on the right.}
\label{apiformat}
\resizebox{\columnwidth}{!}{%
\begin{tabular}{ccccc}
\hline
Data Format                    & PIDF(\%) & \multicolumn{1}{c|}{PODF(\%)} & Conversion Interface        & Description                                   \\ \hline
Scene Layout                & 0        & \multicolumn{1}{c|}{19.05}     & Point\_to\_face\_conversion & Converting points to faces.
            \\
Noise Function              & 0.45     & \multicolumn{1}{c|}{0}         & Cube\_generation            & Generating cubes.                             \\
Image                       & 0.90     & \multicolumn{1}{c|}{0}         & Point\_generation           & Generating points.                            \\
Texture Material            & 0.90     & \multicolumn{1}{c|}{32.54}     & Line\_generation            & Generating lines.                             \\
Geographic Information Data & 1.35     & \multicolumn{1}{c|}{0}         & Face\_generation            & Generating faces.                             \\
Point                       & 3.15     & \multicolumn{1}{c|}{0}         & Line\_to\_face\_conversion  & Converting lines to faces.                    \\
Boolean Value               & 4.50     & \multicolumn{1}{c|}{0}         & Asset\_placement            & Placing assets within a scene or environment. \\
Complex Geometry            & 4.50     & \multicolumn{1}{c|}{43.65}     & Point\_to\_line\_conversion & Converting points to lines.                   \\
Color                       & 5.86     & \multicolumn{1}{c|}{0}         & OSM\_file\_retrieval        & Retrieving OpenStreetMap (OSM) files.         \\
Basic Geometry              & 7.66     & \multicolumn{1}{c|}{0}         & Object\_meshing             & Meshing objects.                              \\
String Information & 8.11  & \multicolumn{1}{c|}{0} & Texture\_information\_extraction       & Extracting texture information.      \\
Surface                     & 12.61    & \multicolumn{1}{c|}{2.38}      & Asset\_material\_retrieval  & Retrieving material data for assets.          \\
Line                        & 13.96    & \multicolumn{1}{c|}{2.38}      & Asset\_mesh\_retrieval      & Retrieving mesh data for assets.              \\
Random Number      & 36.04 & \multicolumn{1}{c|}{0} & Scene\_object\_information\_extraction & Extracting scene object information. \\
                  \hline
\end{tabular}%
}
\end{table}

\begin{figure*}[!tbp]
\centering
\includegraphics[width=1\linewidth]{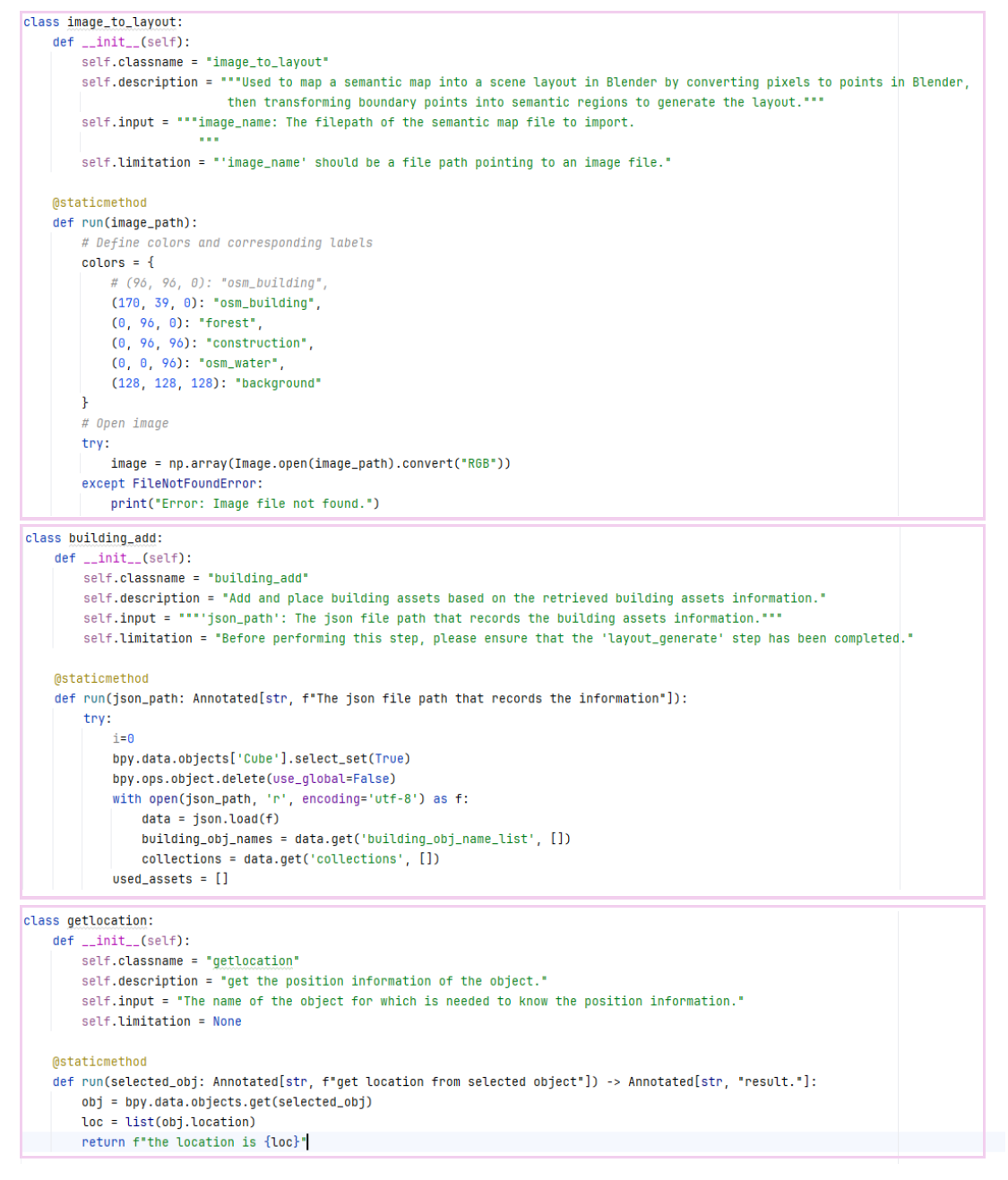}
\caption{Examples of Encapsulated PCG Modules}
\label{fig:daima}
\end{figure*}

\noindent\textbf{Structured Encapsulation.} A crucial component of the protocol is structured encapsulation, for which we have defined specific encapsulation formats. Fig. \ref{fig:daima} illustrates several examples of encapsulated PCG modules, where each section's "Description" also incorporates the mandatory sequential requirements mentioned in Appendix\ref{sec:pcg}. This ensures that multi-agent systems can plan tasks more accurately by referencing the content within the "Description."

\subsection{Infinite Asset Library and Asset Retrieval}
\label{sec:magent}
Although many handcrafted assets are exquisite, in urban scene generation tasks, the compatibility between assets and layout is often more important than the complexity of individual assets. To effectively acquire assets that best match the scene layout, we propose an infinite asset library approach. This includes static assets that we manually collect and annotate, as well as an ever-expanding asset library driven by PCG, providing each asset with text descriptions and image renderings. To accelerate and improve the accuracy of asset retrieval, we use a pre-trained CLIP model to convert text into image retrieval for assets. Each asset's rendered image is encoded into a standardized 768-dimensional vector, which is then compared with the embedding vector of the input description. So far, we have collected and annotated 16,933 static assets and 18 PCGs for generating buildings.The collection of static assets involves annotating properties such as asset paths, footprint dimensions (length, width, and area), center coordinates (with \( z = 0 \)), and asset styles for building assets. Similarly, we record these attributes for PCG-generated buildings. Below is an example JSON structure.

\begin{lstlisting}[language=json]
{
  "buildings": [
    {
      "path": "City\1825\Assets.blend",
      "collection": "Eu_1_building_8",
      "width": 49,
      "length": 38,
      "area": 1862,
      "coordinates": { "x": 158.5, "y": 165 },
      "style": "modern"
    }, ...
    ],
    "plants":[...], ... 
}
\end{lstlisting}

During the building generation process, the following steps are executed:

\begin{itemize}
    \item[1.] \textbf{Bounding Rectangle Calculation} \\
    For each building area \( A \), its minimum bounding rectangle \( R \) is computed to obtain the length \( l \) and width \( w \):
    \begin{multline}
    R = \text{minBoundingRectangle}(A), \\ 
    l = \text{length}(R), \quad w = \text{width}(R)
    \end{multline}

    \item[2.] \textbf{Aspect Ratio Computation} \\
    The aspect ratio of the bounding rectangle is calculated as:
    \begin{equation}
    \text{aspectRatio} = \frac{l}{w}
    \end{equation}

    \item[3.] \textbf{Asset Retrieval and Filtering} \\
    Assets \( \mathcal{S} \) are retrieved from the asset database by matching their footprints and ensuring that their aspect ratio falls within a specified threshold \( \epsilon \):
    \begin{multline}
    \mathcal{S} = \{ s \mid \text{footprint}(s) \approx \text{footprint}(R),\\ \text{aspectRatio}(s) \in [\text{aspectRatio} \pm \epsilon] \}
    \end{multline}
    If assets with an undefined style tag are identified, the CLIP model is employed to filter the asset that best matches the user’s description \( s_{\text{clip}} \).

    \item[4.] \textbf{Scaling Factor Calculation} \\
    For each candidate asset \( s_i \), the scaling factor \( \alpha \) is computed as:
    \begin{equation}
    \alpha_i = \frac{\min(l, w)}{\min(l_i, w_i)}
    \end{equation}
    The final target building is determined by selecting the asset with the smallest scaling factor:
    \begin{equation}
    s_{\text{target}} = \arg\min_{s_i \in \mathcal{S}} \alpha_i
    \end{equation}

    \item[5.] \textbf{Position Offset and Rotation} \\ 
    After selecting the target asset \( s_{\text{target}} \), its position is adjusted based on the asset’s center coordinates \( (x_{\text{center}}, y_{\text{center}}) \) and placed at the appropriate location within the scene. The rotation angle \( \theta \) of the target asset is determined by the angular offset of the bounding rectangle:  
    \begin{equation}
    \theta = \text{angle}(R) - \text{angle}(s_{\text{target}})
    \end{equation} 
    Where \( \text{angle}(R) \) is the angle of the bounding rectangle, and \( \text{angle}(s_{\text{target}}) \) is the original rotation angle of the target asset. The final rotation transformation for the target asset is:  
    \begin{equation}
    s_{\text{target}} = \text{rotate}(s_{\text{target}}, \theta)
    \end{equation}  
    The asset’s position is then adjusted with an offset \( \Delta x, \Delta y \):  
    \begin{equation}
    s_{\text{target}} = \text{translate}(s_{\text{target}}, \Delta x, \Delta y)
    \end{equation}  
\end{itemize}

\begin{figure*}[!tbp]
\centering
\includegraphics[width=1\linewidth]{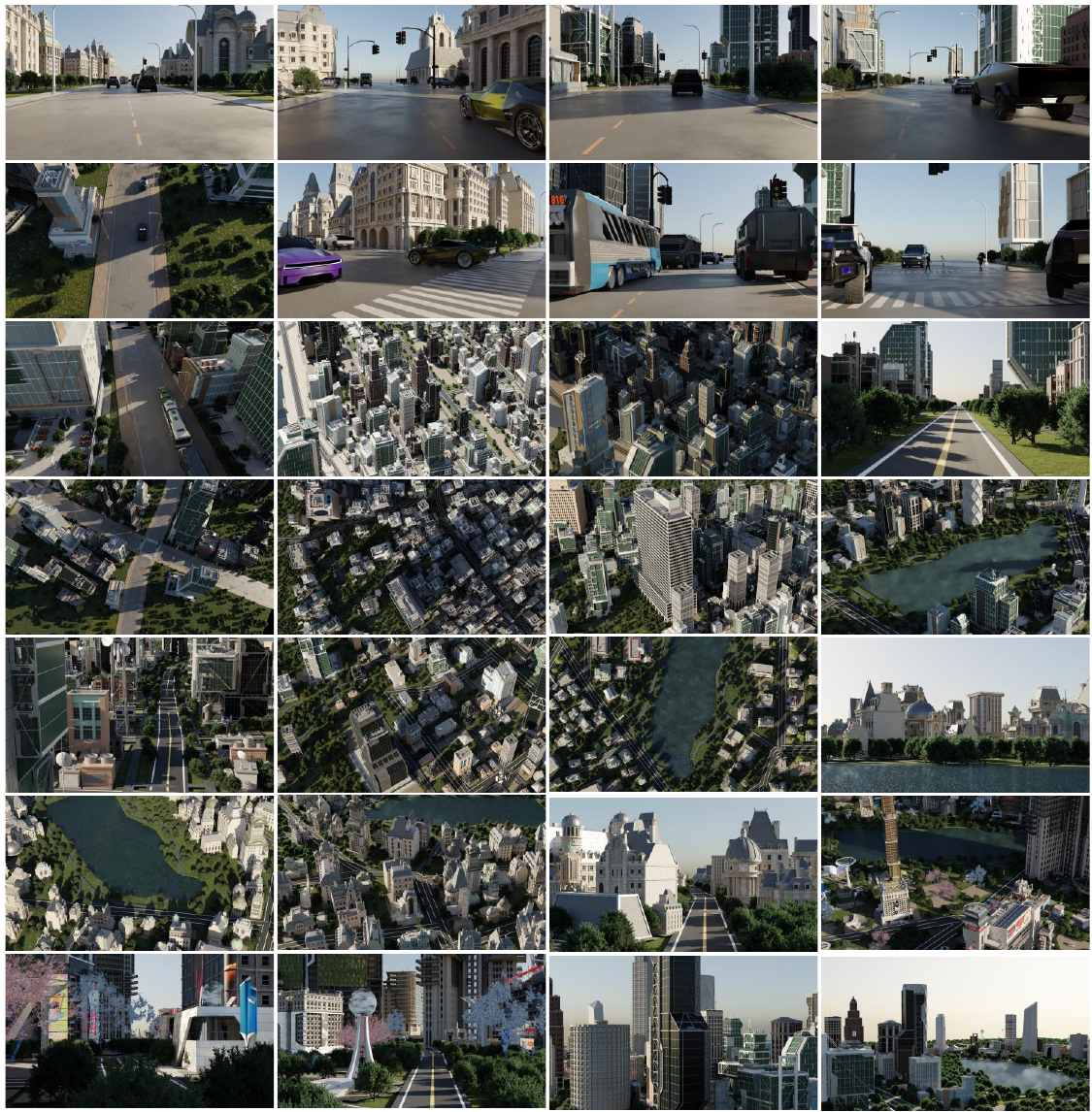}
\caption{Random Outputs from the Dataset Generator.}
\label{fig:datamore}
\end{figure*} 

\begin{table*}[]
\caption{Comparison of \mymethod{} Synthetic Dataset with Existing Urban Datasets.}
\label{tab:dataset}
\small
\begin{tabular}{llccccccc}
\hline
Dataset          & Year & \begin{tabular}[c]{@{}c@{}}Scenes \\ in Total\end{tabular} & \begin{tabular}[c]{@{}c@{}}Assets \\ in Total\end{tabular} & \begin{tabular}[c]{@{}c@{}}Area/Length \\ in Total\end{tabular} & \begin{tabular}[c]{@{}c@{}}Triangles \\ Pre-Scene\end{tabular} & \begin{tabular}[c]{@{}c@{}}Free \\ Assets\end{tabular} & Annotation & \begin{tabular}[c]{@{}c@{}}Provides \\ Code\end{tabular} \\ \hline
Semantic3D\cite{Hackel_Savinov_Ladicky_Wegner_Schindler_Pollefeys_2017}       & 2017 & 3                                                          & -                                                          & -                                                               & 4000M                                                          & No          &  Semantic           &  No             \\
Synscapes\cite{Wrenninge_Unger_2018}        & 2018 & 25K                                                     & -                                                          & -                                                               & -                                                              & No          &   Semantic          & No            \\
DublinCity\cite{Zolanvari_Ruano_Rana_Cummins_Silva_Rahbar_Smolic_2019}       & 2019 & 1                                                          & -                                                           & 2$ \text{km}^2 $                                                             &   260M                                                             & No          &  Semantic           &  No             \\
ProcSy\cite{Khan_Phan_Salay_Czarnecki_2019}           & 2019 & $\infty$                                                       & $\infty$                                                           &  $\infty$                                                               & -                                                              & Yes         & Semantic           & No            \\
Campus3D\cite{Li_Li_Tong_Lim_Yuan_Wu_Tang_Huang_2020}         & 2020 & 1                                                          &  -                                                          &  1.58$ \text{km}^2 $                                                              & 937.1M                                                               & No          & Hierarchical          &  No             \\
Meta-Sim2\cite{Devaranjan_Kar_Fidler_2020}        & 2020 & $\infty$                                                       &  -                                                          &   -                                                              & -                                                              & No          & Semantic           & No            \\
Hessigheim 3D\cite{Kölle_Laupheimer_Schmohl_Haala_Rottensteiner_Wegner_Ledoux_2021}    & 2021 & 1                                                          &  -                                                          & 0.19 $ \text{km}^2 $                                                      & 125.7M/36.76M                                                               & No          &Semantic            & No              \\
SUM\cite{Gao_Nan_Boom_Ledoux_2021}              & 2021 & 1                                                          & -                                                           & 4  $ \text{km}^2 $                                                              & 19M                                                               & No          & Semantic           & No              \\
STPLS3D\cite{Chen_Hu_Hugues_Feng_Hou_McCullough_Soibelman}          & 2022 & 1                                                          & -                                                           &  1.27$ \text{km}^2 $                                                                & 150.4M                                                               & No          &Semantic            & No              \\
InstanceBuilding\cite{Xu_Chen_Lu_Liang_Nan} & 2022 & 1                                                          & -                                                           & 0.434$ \text{km}^2 $                                                                 & 7.46M                                                               & No          &Semantic            &  No             \\
UrbanBIS\cite{Yang_Xue_Zhang_Xie_Fu_Huang_2023}         & 2023 & 5                                                          & -                                                           & 10.78$ \text{km}^2 $                                                                 &  2523.8M/284.3M                                                        & No          & Semantic/Instance           & No              \\
MatrixCity\cite{Li_Jiang_Xu_Xiangli_Wang_Lin_Dai_2023}       & 2024 & 2                                                          & -                                                           & 28$ \text{km}^2 $                                                                &  -                                                              & Yes         &Semantic/Instance/Depth            & Yes              \\
\textbf{CityX(Ours)}      & 2024 & $\infty$                                                       & $\infty$                                                           & $\infty$                                                                &  383M/$ \text{km}^2 $                                                              & Yes         & Semantic/Instance/Depth  & Yes             \\
        \hline
\end{tabular}
\end{table*}

\begin{figure*}[!tbp]
\centering
\includegraphics[width=1\linewidth]{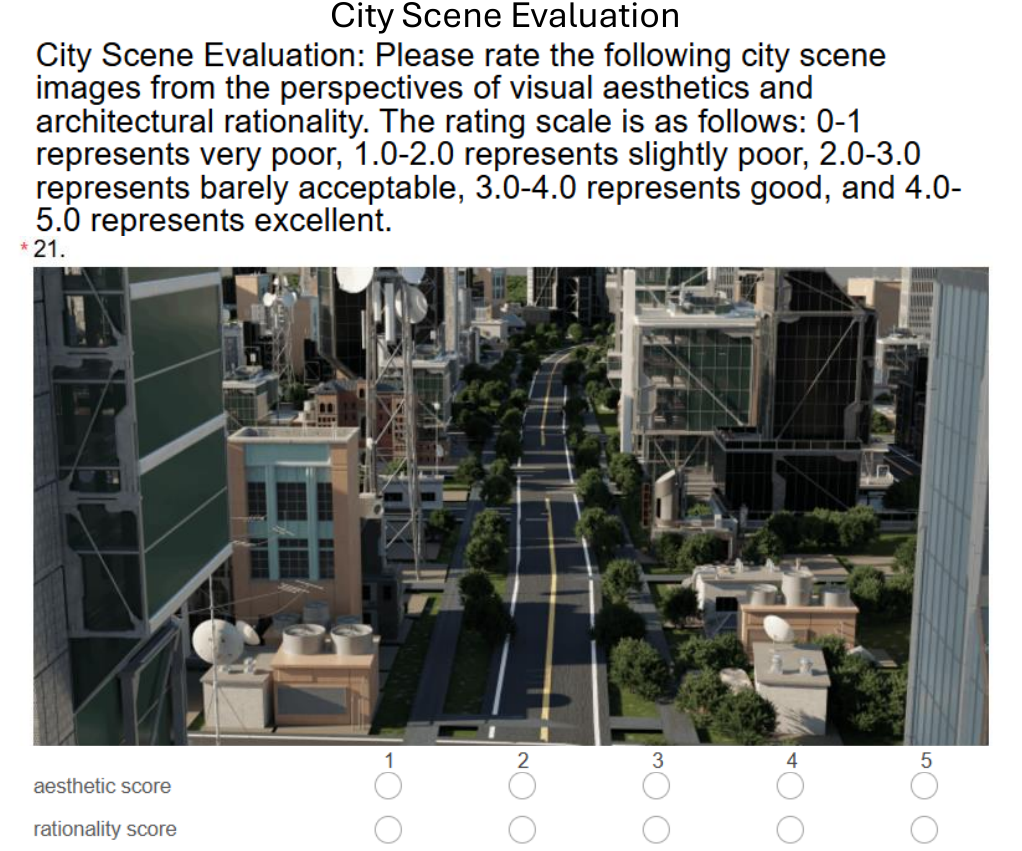}
\caption{The example questionnaire for participants.}
\vspace{-0.5cm}
\label{fig:as}
\end{figure*} 

\subsection{Details of Multi-agent}
\label{sec:magent}

The Annotator(mentioned in Sec \ref{sec:agent}) labels action functions in two steps: first, summarizing existing functions into consistent concepts guided by prompts; second, labeling each function based on these concepts. An action function may receive multiple labels. Once all action functions are processed, the labeled information is stored in the common message pool, enabling other agents to directly access it. 

\section{Dataset Synthetic}
\label{sec:data}

Synthetic data for urban-scale scenes is widely used in many tasks within computer vision, as real-world data collection incurs significant costs, and manual modeling faces similar challenges. Our  method \mymethod{} generates synthetic datasets with high quality and efficiency while requiring minimal resources. Tab. \ref{tab:dataset} compares the \mymethod{} synthetic dataset with several other urban datasets. We provide a large sample of random RGB images from our dataset generator(Fig. \ref{fig:datamore}), each image was rendered simultaneously for a variety of channels (various specification images for the dataset in Sec \ref{sec:dataset}. 

\section{User Study Details}
\label{sec:study}

In our experimental process, we employed manual evaluation to assess the results. An example of the evaluation interface used during this process is shown in Fig. \ref{fig:as}. To ensure fairness and objectivity, all evaluation images were anonymized to prevent bias in the assessment. This approach maintained the integrity of the evaluation process, ensuring that each image was judged solely based on its merits.

{
    \small
    \bibliographystyle{ieeenat_fullname}
    \bibliography{main}
}



\end{document}